\documentclass[wcp]{jmlr}
\usepackage{amsmath,amssymb,graphicx,url}
\jmlrvolume{Pages}
\jmlryear{2019}
\jmlrworkshop{}
\jmlrproceedings{}{}

\usepackage[ruled]{algorithm}
\usepackage{amsmath}
\usepackage{varwidth}
\usepackage{amssymb}
\usepackage{mathtools, cuted}
\usepackage{booktabs}
\usepackage{graphicx}
\usepackage{float}

\usepackage{forest}
\usetikzlibrary{shadows}

\usepackage[shortlabels]{enumitem}
\begin{document}

\title[NDCP Regression]{Combining Prediction Intervals on Multi-Source Non-Disclosed Regression Datasets}

\author{\Name{Ola Spjuth}\Email{ola.spjuth@farmbio.uu.se}\\
       \addr{Department of Pharmaceutical Biosciences
       Uppsala University, Uppsala, Sweden}\\
       \Name{Robin Carri\'{o}n Br\"{a}nnstr\"{o}m}\Email{robincarrion@gmail.com} \\
       \addr{Department of Statistics,  Uppsala University, Uppsala, Sweden}\\
       \Name{Lars Carlsson}\Email{lars.carlsson@stenaline.com}\\
       \addr{Stena Line AB, Gothenburg, Sweden}\\
        \Name{Niharika Gauraha}\Email{niharika.gauraha@farmbio.uu.se}\\
       \addr{Department of Pharmaceutical Biosciences
       Uppsala University, Uppsala, Sweden}
      }


\maketitle

\begin{abstract}
Conformal Prediction is a framework that produces prediction intervals based on the output from a machine learning algorithm. In this paper we explore the case when training data is made up of multiple parts available in different sources that cannot be pooled. We here consider the regression case and propose a method where a conformal predictor is trained on each data source independently, and where the prediction intervals are then combined into a single interval. We call the approach Non-Disclosed Conformal Prediction (NDCP), and we evaluate it on a regression dataset from the UCI machine learning repository using support vector regression as the underlying machine learning algorithm, with varying number of data sources and sizes. The results show that the proposed method produces conservatively valid prediction intervals, and while we cannot retain the same efficiency as when all data is used, efficiency is improved through the proposed approach as compared to predicting using a single arbitrarily chosen source.
\end{abstract}

\section{Introduction}
There is a growing number of data analysis applications in which data comes from multiple unrelated sources and full disclosure of the data between the parties is prevented by privacy and security concerns~\citep{abadi2017protection,Papernot2018}.
In such predictive analysis settings, the challenge is to make use of the isolated data sources in statistical learning systems, with the objective to make more accurate predictions on future objects without sharing the data with other sources. 
Pooling of data to one particular location for model building can be a possible solution for small data sources, especially when data privacy is not a concern. However, if data is large or if the data owners do not allow such pooling of data, one has to resort to secure and distributed learning approaches such as secure federated learning methods.
However, federated learning models, (for example \cite{Shokri:2015os} for deep learning) are usually  complex to implement in practice.


We present a light-weight framework that gives more accurate prediction intervals 
by aggregating conformal predictions (prediction intervals) computed at individual locations (data sources)
without sharing the data between the sources. Conformal Prediction is a framework that complements the prediction from a machine learning algorithm with a valid measure of confidence (i.e. prediction intervals) assuming that the data is exchangeable~\citep{vovk2005algorithmic}. 
We propose to combine conformal predictions from multiple sources, where inductive conformal predictors~\citep{papadopoulos2002inductive} and cross conformal predictors~\citep{vovk2015cross} are applied on the multiple data sources and their individual prediction intervals are combined to form a single prediction on a new example. We refer to this method as Non-Disclosed Conformal Prediction (NDCP).

The organization of the paper is as follows. In section 2, we introduce the background concepts and notations used throughout the paper. In Section 3, we introduce the concept of aggregating conformal predictions from multiple sources. The experimental setup and experiments are described in Section 4. 
A discussion is presented in Section 5, and the paper is concluded in Section 6.

\section{Background}
In this paper, we consider only regression problems and assume exchangeability of observations. The object space is denoted by $\mathcal{X} \subset \mathbb{R}^p$, where $p$ is the number of features, and  label space is denoted by $\mathcal{Y} \subset \mathbb{R}$. We assume that each example consists of an object and its label, and its space is given as $\mathcal{Z} := \mathcal{X} \times \mathcal{Y}$. 
In a classical regression setting, given $\ell$ data points $Z = \{ z_1 , ..., z_\ell \}$  where each example $z_i = (x_i, y_i)$ is labeled, we want to predict the label of a new object $x_{new}$. 

In the conformal prediction setting, a non-conformity measure is the score from a function that measures the strangeness of an example in relation to the previous examples~\citep{vovk2005algorithmic}. For regression problems, a commonly used non-conformity measure is
\begin{align}\label{eq:alpha1}
    \alpha_i = |y_i - \hat{y_i}|,
\end{align}

\noindent where $\hat{y}_i$ is the estimated output for the object $x_i$ using regression algorithms. When using the non-conformity measure (\ref{eq:alpha1}), the prediction intervals will be of equal length for all test examples. Instead, a non-conformity measure which takes into account the accuracy of the decision rule $f$ on $x_i$ can be used, yielding a prediction interval with a length proportional to the predicted accuracy of the new example. i.e., the prediction intervals will be tighter when the underlying algorithm's prediction is good and larger when it is predicted to be bad. The normalized non-conformity score is
\begin{align}\label{eq:alpha2}
   \alpha_i= \left|\frac{y_i - \hat{y}_i}{\exp (\sigma_i)} \right|,
\end{align}
where $\sigma_i$ is the prediction of the logarithm of the absolute residuals, $\ln(y_i - \hat{y}_i)$, from a linear SVR trained on the proper training set. This is considered to be an estimate of the decision rule accuracy \citep{papadopoulos2002inductive}.

Conformal predictors are built on top of standard machine learning algorithms and complement the predictions with valid measures of confidence~\citep{vovk2005algorithmic}. The two main approaches are Transductive Conformal Prediction (TCP)~\citep{vovk2013transductive} and Inductive Conformal Prediction (ICP)~\citep{papadopoulos2002inductive} and they can be used for both classification and regression problems. TCP is computationally demanding; for every test example a re-training of the model is required, and ICP was developed to overcome this issue. In ICP, a subset of training examples are set aside for calibration which makes it less informational efficient. To address this problem of information efficiency, ensembles of conformal predictors were introduced such as Cross Conformal Prediction (CCP)~\citep{vovk2015cross,papadopoulos2015cross}, Aggregated Conformal Prediction (ACP)~\citep{carlsson2014aggregated}, Combination of inductive mondrian conformal predictors~\citep{toccaceli2018combination} etc. These ensemble methods aim to construct more informational efficient conformal predictors by combining p-values. However, most of the resulting models are not guaranteed to be valid, as the combined p-values need not be uniformly distributed~\citep{Linusson:2017dn}. Also, various methods of combining p-values have been proposed, for example, combining p-values using their mean~\citep{vovk2015cross}, using their median~\citep{Linusson:2017dn}, using extended  chi-square  function and using standard  normal  form~\citep{balasubramanian2015conformal}.



\section{Non-Disclosed Conformal Prediction}
In this section, we present the proposed method which we call Non-Disclosed Conformal Prediction (NDCP). Mainly, we propose a new framework to combine conformal prediction (CP) intervals across various data sources where the number of sources, the size of each data source and the distribution of data may vary, and where data is not shared between the data sources. 

Suppose we have $K$ data sources, each with a training dataset $D_k$ of arbitrary sizes where $k \in {1,...,K}$. For a new object $x_{new}$, the objective is to combine prediction intervals at the location $A$ that were computed in each data source using CP. The result is a set of aggregated prediction-intervals, where no training data is disclosed between the data sources and between the data sources and location $A$, but the only information that is transmitted between data sources and $A$ is the object to predict and the resulting prediction-intervals. 
Assuming that data owners can not disclose anything else than point predictions and intervals, a simple and relatively naive approach is to take the average of all shared values. The intervals are combined by using the median upper and lower bound respectively. An overview of the NDCP algorithm is presented in Algorithm \ref{algo:NDCP}.

\vspace{1cm}

 \begin{algorithm}[H]
 \caption{Non-Disclosed Conformal Prediction (NDCP)} \label{algo:NDCP}
 \textbf{Input:}{ $K$ Data sources: $D_1,...,D_K $, test example: $x_{new}$}\\
 \textbf{Output:}{\textbf{ The prediction interval for $x_{new}$: $I$} }\\
 \textbf{Steps:}\\
 \For{each $D_k$,  $k \in \{ 1,...,K\} $ }{
    Train an ICP or CCP and compute prediction interval $I_k$ for the test example $x_{new}$\\
	Transfer $I_k$ to location $A$\\
 }
 Combine all intervals into one interval $I$ (by taking the medians of the lower and upper bounds) at location $A$
  
 \textbf{return} $I$
 \end{algorithm}

\section{Experiments}
We evaluate NDCP on the benchmark data set Concrete Compressive Strength from the UCI repository~\citep{lichman2013uci}. The experimental setup and experiments are described in the following subsections.

\subsection{Experimental Setup}
To simulate a scenario where data are located in different places, data is split into subsets where each subset represents an individual data source. After a test set has been set aside, data is split in three different ways to simulate different scenarios:
\begin{enumerate}
	\item Equally sized data sources: Training set is randomly partitioned into equally sized data sources.
	\item Unequally sized data sources: Training set is randomly partitioned into different sizes, simulating a real life scenario where one data source is larger than the rest.
	\item Non-IID, equally sized data sources: Training set is divided such that one of the data sources has higher proportion of observation with high values of the response variable, simulating a real life scenario where different data owners do not have identical data.
\end{enumerate}

The evaluation procedure is outlined below:

\begin{enumerate}
	\item Randomly split the data set into a training set (90\%) and a test set (10\%)
	\item Split the training set into $ K $ disjoint data sets of
	\begin{enumerate}[(a),noitemsep]
		\item random equally sized data sources
		\item random unequally sized data sources
		\item non-IID, equally sized data sources
	\end{enumerate}
	\item Train ICP or CCP on each individual data set 
	\item Aggregate predictions from all $ K $ data sets using NDCP
	\item Train ICP or CCP on the pooled data from all $ K $ data sets
	\item Repeat step one to five 100 times
\end{enumerate}

In our experiments, we use the following as the underlying machine algorithms: Support Vector Regression (SVR) with an RBF kernel, a linear SVR as proposed in \cite{papadopoulos2015cross}, and Random Forests~\citep{Breiman}. The nonconformity measures were calculated as given in equation (\ref{eq:alpha1}). The prediction intervals were combined by taking the medians of the lower and upper bounds as suggested in~\cite{park2015aggregating}.

For evaluations, we consider validity and efficiency. Validity is the proportion of true values contained in the prediction interval. As efficiency metric, we use the median width of prediction intervals. $n$ represents the total number of observations in the training set for each data source. Note that $n$ for NDCP refers to the sum of all observations used in all the models producing the prediction intervals that are combined, and is hence the total number of observations (referred to as Pooled). The objective of NDCP is primarily to have improved efficiency compared to the individual sources that are used to create the NDCP intervals. It is also desirable that the performance is as close to the pooled data as possible.

Together with the results, a hypothetical \emph{Ideal NDCP} is also presented. This represents an NDCP with an ideal combination of intervals, in the sense that an exact validity is attained. I.e. if the intervals are conservative (predicted error is less than expected error), the intervals will be shrunk symmetrically with the same factor until the expected error rate is obtained. Note that this is only possible to do after the true labels have been revealed, and is only included to show the results NDCP would give with an hypothetical, optimal symmetric interval combination.

Each setting is considered with 2, 4 and 6 different data sources and is repeated 100 times to obtain consistent results. Support Vector Regression (SVR) with an RBF kernel was used and in every run the parameters $C$, $\varepsilon$ and $\gamma$ were optimized through grid search and selected through 10-fold cross-validation. When creating prediction intervals a significance level of 5\% is used.

\subsection{Experimental Results}\label{sec:results}
This section is divided into three parts, investigating each of the three different settings for splitting the data.

\subsubsection{Experiment 1: Equally sized data sources}
Results from splitting the data into 2, 4, and 6 equally sized data sources are presented in Table~\ref{tab:res_equal}. Each row in the table represents the results from one specific model or data source. At the 5\% confidence level we observe that NDCP using ICP in all cases has a lower efficiency when compared to the individual data sources, except for 2 data sources but here NDCP has lower efficiency than one of the data sources. 


\begin{table}[!ht]
	\centering
		\footnotesize
	\caption{Results from Experiment 1, Equally sized data sources, for models NDCP, Ideal NDCP, the individual equally sized data sources (2, 4 and 6) and Pooled. Results for validity (Val) and prediction interval median width as a measure of efficiency (Eff) are listed in the columns for CCP and ICP at different confidence levels. The column $n$ refers to the number of observations underlying the predictions.} \label{tab:res_equal}
	\begin{tabular}{lccccccccccc}
		\toprule
	&	& \multicolumn{2}{c}{CCP 5\%} & \multicolumn{2}{c}{ICP 5\%}& \multicolumn{2}{c}{ICP 10\%}& \multicolumn{2}{c}{ICP 15\%}& \multicolumn{2}{c}{ICP 20\%} \\
\textbf{2 sources}	& \textbf{n} & \textbf{Val}	& \textbf{Eff} & \textbf{Val}	& \textbf{Eff} & \textbf{Val}	& \textbf{Eff} & \textbf{Val} & \textbf{Eff} & \textbf{Val} & \textbf{Eff}  \\ 
		\midrule
NDCP & 927 & 0.971 & 26.813 & 0.966 & 28.566 & 0.925 & 22.391 & 0.882 & 18.700 & 0.823 & 16.266\\
Ideal NDCP & 927 & 0.950 & 24.575 & 0.950 & 26.717 & 0.900 & 20.961 & 0.850 & 17.392 & 0.800 & 15.497\\
Source1 & 463 & 0.957 & 26.965 & 0.945 & 29.155 & 0.891 & 22.364 & 0.848 & 18.804 & 0.789 & 16.100\\
Source2 & 464 & 0.960 & 26.748 & 0.940 & 28.051 & 0.897 & 22.484 & 0.847 & 18.709 & 0.794 & 16.426\\
Pooled & 927 & 0.964 & 22.283 & 0.945 & 23.259 & 0.900 & 18.445 & 0.854 & 15.362 & 0.794 & 13.286\\

\toprule
\textbf{4 sources}& \\
\midrule
NDCP & 927 & 0.978 & 31.674 & 0.967 & 31.738 & 0.936 & 26.257 & 0.881 & 21.596 & 0.851 & 19.379\\
Ideal NDCP & 927 & 0.950 & 27.630 & 0.950 & 29.291 & 0.900 & 23.688 & 0.850 & 20.159 & 0.800 & 17.442\\
Source1 & 231 & 0.958 & 31.655 & 0.925 & 32.020 & 0.883 & 27.039 & 0.819 & 21.523 & 0.791 & 19.393\\
Source2 & 232 & 0.958 & 31.858 & 0.921 & 32.803 & 0.882 & 26.037 & 0.826 & 21.733 & 0.793 & 19.702\\
Source3 & 232 & 0.955 & 31.793 & 0.928 & 32.325 & 0.887 & 26.804 & 0.830 & 22.026 & 0.781 & 19.350\\
Source4 & 232 & 0.957 & 32.186 & 0.930 & 33.484 & 0.895 & 26.650 & 0.829 & 22.127 & 0.794 & 19.552\\
Pooled & 927 & 0.962 & 22.052 & 0.945 & 23.474 & 0.895 & 18.449 & 0.845 & 15.524 & 0.791 & 13.488\\

		\toprule
\textbf{6 sources}&  \\
\midrule
NDCP & 927 & 0.977 & 34.492 & 0.964 & 33.214 & 0.937 & 27.859 & 0.897 & 24.080 & 0.841 & 20.463\\
Ideal NDCP & 927 & 0.950 & 29.853 & 0.950 & 31.196 & 0.900 & 25.119 & 0.850 & 21.653 & 0.800 & 18.911\\
Source1 & 155 & 0.952 & 35.010 & 0.921 & 34.886 & 0.876 & 28.786 & 0.822 & 24.234 & 0.767 & 20.520\\
Source2 & 155 & 0.953 & 34.711 & 0.914 & 36.682 & 0.876 & 28.211 & 0.827 & 24.757 & 0.774 & 20.741\\
Source3 & 154 & 0.952 & 35.229 & 0.912 & 35.570 & 0.881 & 30.190 & 0.834 & 24.790 & 0.779 & 21.007\\
Source4 & 154 & 0.953 & 34.876 & 0.914 & 36.096 & 0.861 & 28.441 & 0.830 & 24.333 & 0.765 & 20.491\\
Source5 & 154 & 0.952 & 34.525 & 0.913 & 33.503 & 0.881 & 28.846 & 0.827 & 24.578 & 0.774 & 20.462\\
Source6 & 155 & 0.949 & 34.547 & 0.914 & 35.850 & 0.877 & 28.149 & 0.834 & 24.676 & 0.769 & 20.322\\
Pooled & 927 & 0.967 & 22.601 & 0.948 & 22.932 & 0.897 & 18.490 & 0.845 & 15.440 & 0.800 & 13.366\\
\bottomrule
\end{tabular}
\end{table}

\begin{figure}[t]
  {%
    \subfigure[ICP]{\label{fig:boxplot_equal_icp}%
      \includegraphics[width=0.45\linewidth]{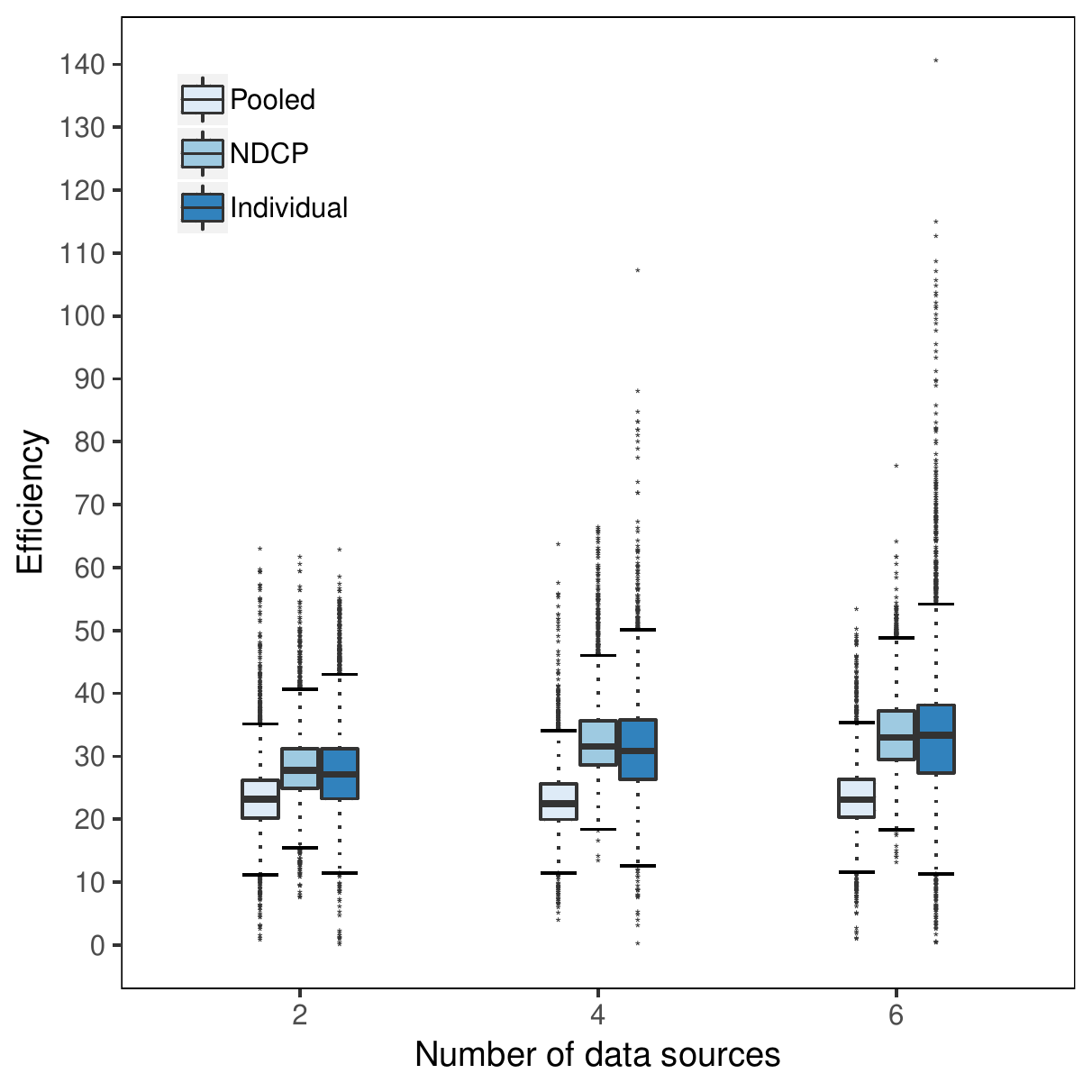}}%
    \qquad
    \subfigure[CCP]{\label{fig:boxplot_equal_ccp}%
      \includegraphics[width=0.45\linewidth]{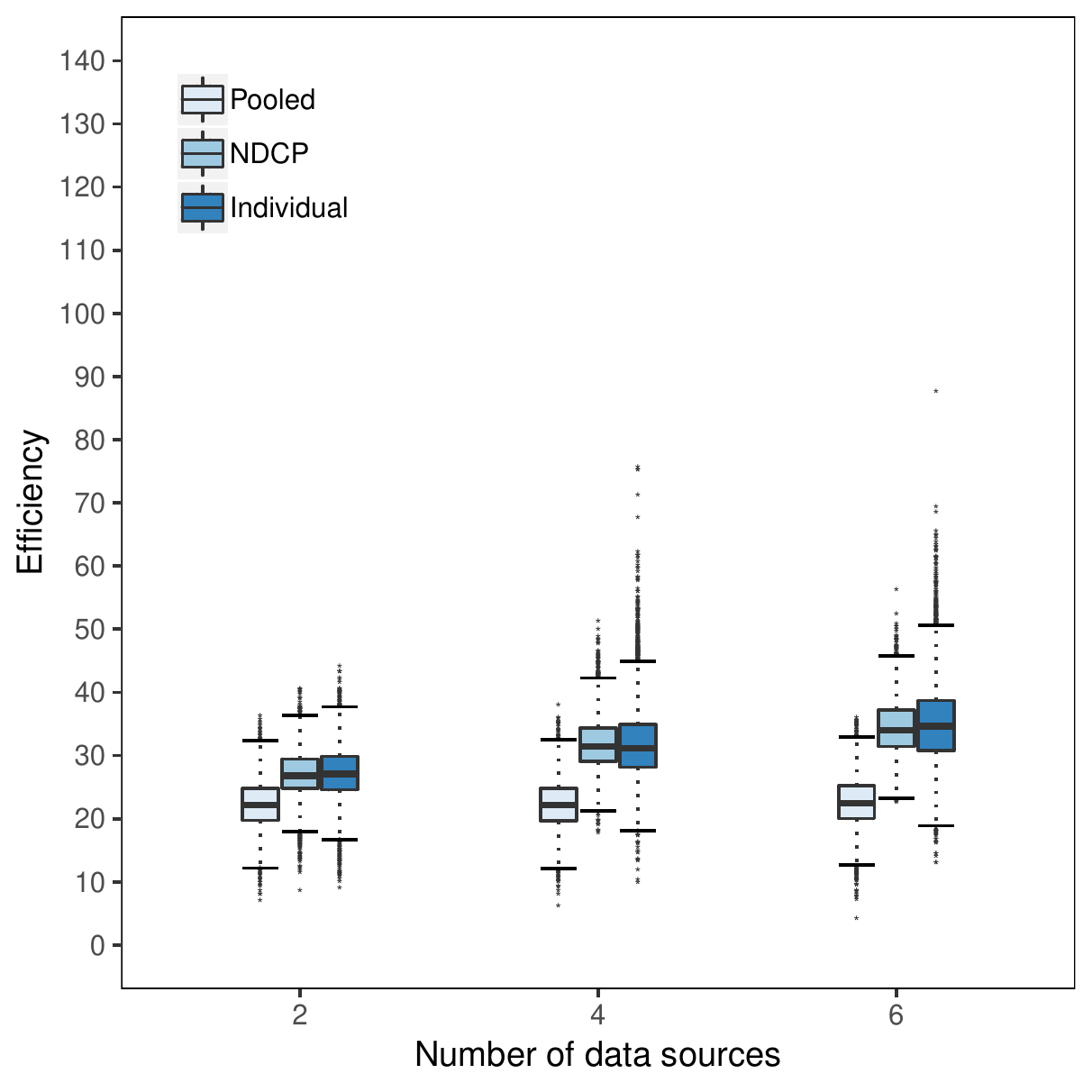}}
  }
 \caption{Dispersion of efficiency (prediction interval widths) for Experiment 1; equally sized data sources, for 2, 4 and 6 number of data sources for ICP and CCP. Results from Pooled, NDCP and a randomly selected data source (named Individual) from the equally sized data sources are presented.}
  \label{fig:boxplot_equal}
\end{figure}

In Figure \ref{fig:boxplot_equal} the dispersion of the prediction interval widths are presented for Pooled, NDCP and a randomly selected data source from the equally sized data sources where CCP is used for each model at 5\% confidence level. We observe that the prediction interval widths coming from the individual sources always has larger variance as compared to NDCP.


\subsubsection{Experiment 2: Unequally sized data sources}

In Table \ref{tab:res_size} the results from splitting the data into unequal sources are presented so that Source1 contains approximately twice as many observations as the smaller sources. 
As expected, the larger Source1 in all cases has lower efficiency than the smaller Source2. We also note that NDCP in all cases has a lower efficiency than at least one of the individual data sources, this is true for both ICP and CCP. NDCP also improves in terms of validity compared with the smaller data sources.
Using the hypothetical, optimal combination of intervals with Ideal NDCP, interval widths approach the same values as of the larger Source1.

\begin{table}
	\centering
	\footnotesize
	\caption{Results from Experiment 2, Unequally sized data sources, for models NDCP, Ideal NDCP, the individual equally sized data sources (2, 4 and 6) and Pooled. All models are listed in the first column and the corresponding  validity and PI median width as a measure of efficiency are listed in the next four columns for CCP and ICP respectively. $n$ refers to the number of observations underlying the predictions.} 
	\label{tab:res_size}
	\begin{tabular}{lcccccc}
		\toprule
		& \multicolumn{2}{c}{CCP} & \multicolumn{2}{c}{ICP} \\
		\textbf{2 sources} & \textbf{Validity} & \textbf{Efficiency} & \textbf{Validity} & \textbf{Efficiency} & \textbf{n} \\ 
		\midrule
		NDCP & 0.971 & 27.453 & 0.961 & 28.438 & 927 \\ 
		Ideal NDCP   & 0.950 & 24.921 & 0.950 & 26.928 & 927 \\
		Source1 &  0.957 & 25.064 & 0.942 &25.990 & 620 \\ 
		Source2   & 0.957 & 29.951 & 0.938 &30.899 & 307 \\ 
		Pooled   & 0.963 & 22.221 & 0.943 &22.962  & 927 \\ 
		\midrule
		\textbf{4 sources}& \\
		\midrule
		NDCP  & 0.973 & 30.472 & 0.971 & 32.127 & 927 \\ 
		Ideal NDCP   & 0.950 & 27.336 & 0.950 & 29.188  & 927 \\
		Source1   & 0.957 & 28.534 & 0.940 & 29.350 & 370 \\ 
		Source2   & 0.946 & 31.622 & 0.930 & 32.902 & 185 \\ 
		Source3   & 0.949 & 31.815 & 0.932 & 35.257  & 186 \\ 
		Source4   & 0.935 & 30.782 & 0.929 & 36.080 & 186 \\ 
		Pooled  & 0.963 & 22.106 & 0.945 & 23.252 & 927 \\ 
		\midrule
		\textbf{6 sources}&\\ 
		\midrule
		NDCP  & 0.973 & 32.986 & 0.972 & 35.562  & 927 \\ 
		Ideal NDCP   & 0.950 & 29.489 & 0.950 & 31.491 & 927 \\
		Source1  & 0.947 & 29.295 & 0.933 & 31.743 & 260 \\ 
		Source2   & 0.940 & 33.643 & 0.927 & 39.362 & 134 \\ 
		Source3  & 0.945 & 34.811 & 0.926 & 36.024 & 134 \\ 
		Source4   & 0.941 & 34.251 & 0.925 & 44.878 & 133 \\ 
		Source5   & 0.944 & 35.056 & 0.931 & 42.282 & 133 \\ 
		Source6   & 0.940 & 33.647 & 0.926 & 44.392 & 133 \\ 
		Pooled   & 0.963 & 22.128 & 0.948 & 23.330 & 927 \\ 
		\bottomrule
	\end{tabular}
\end{table}

\begin{figure}[htbp]
  {%
    \subfigure[ICP]{\label{fig:boxplot_size_icp}%
      \includegraphics[width=0.45\linewidth]{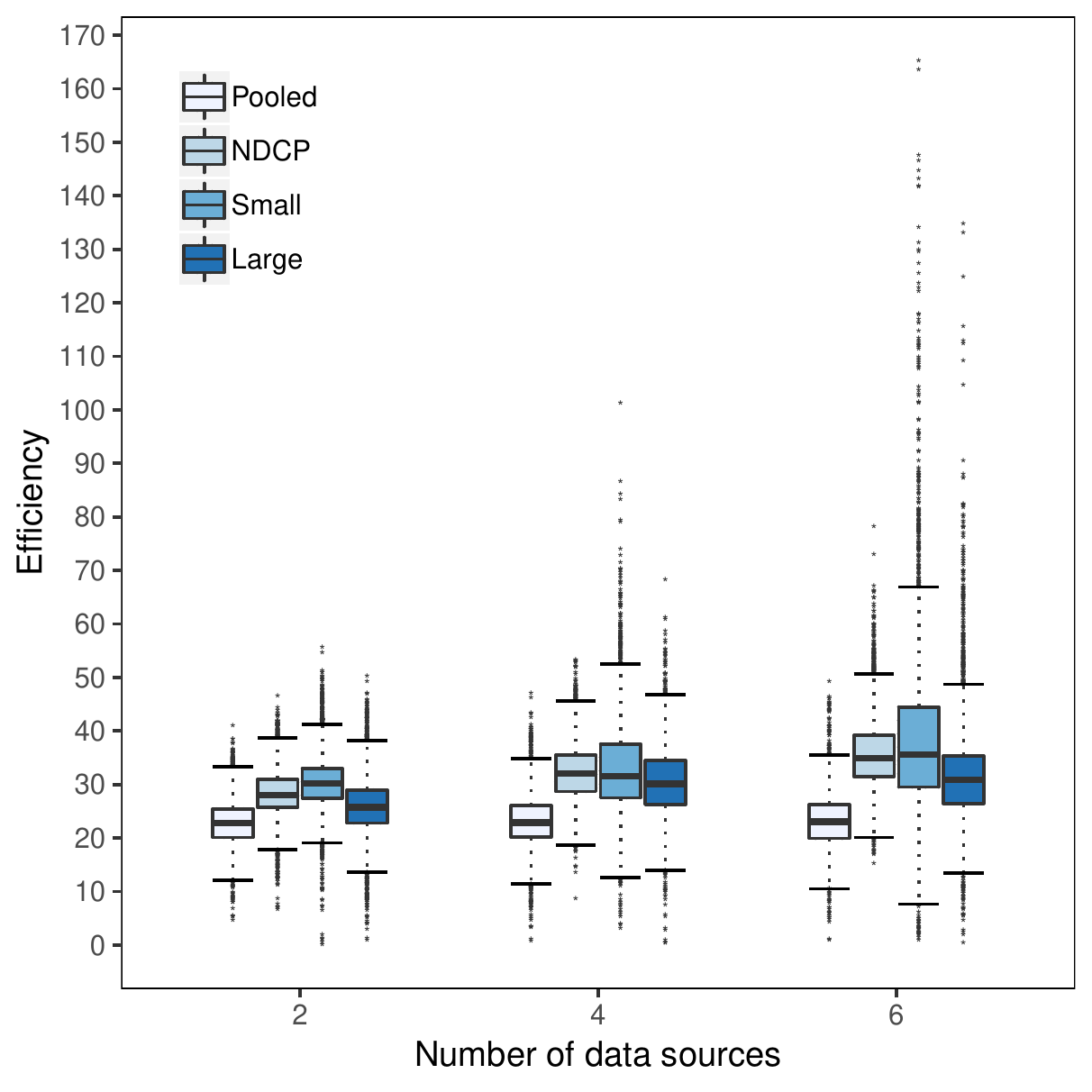}}%
    \qquad
    \subfigure[CCP]{\label{fig:boxplot_size_ccp}%
      \includegraphics[width=0.45\linewidth]{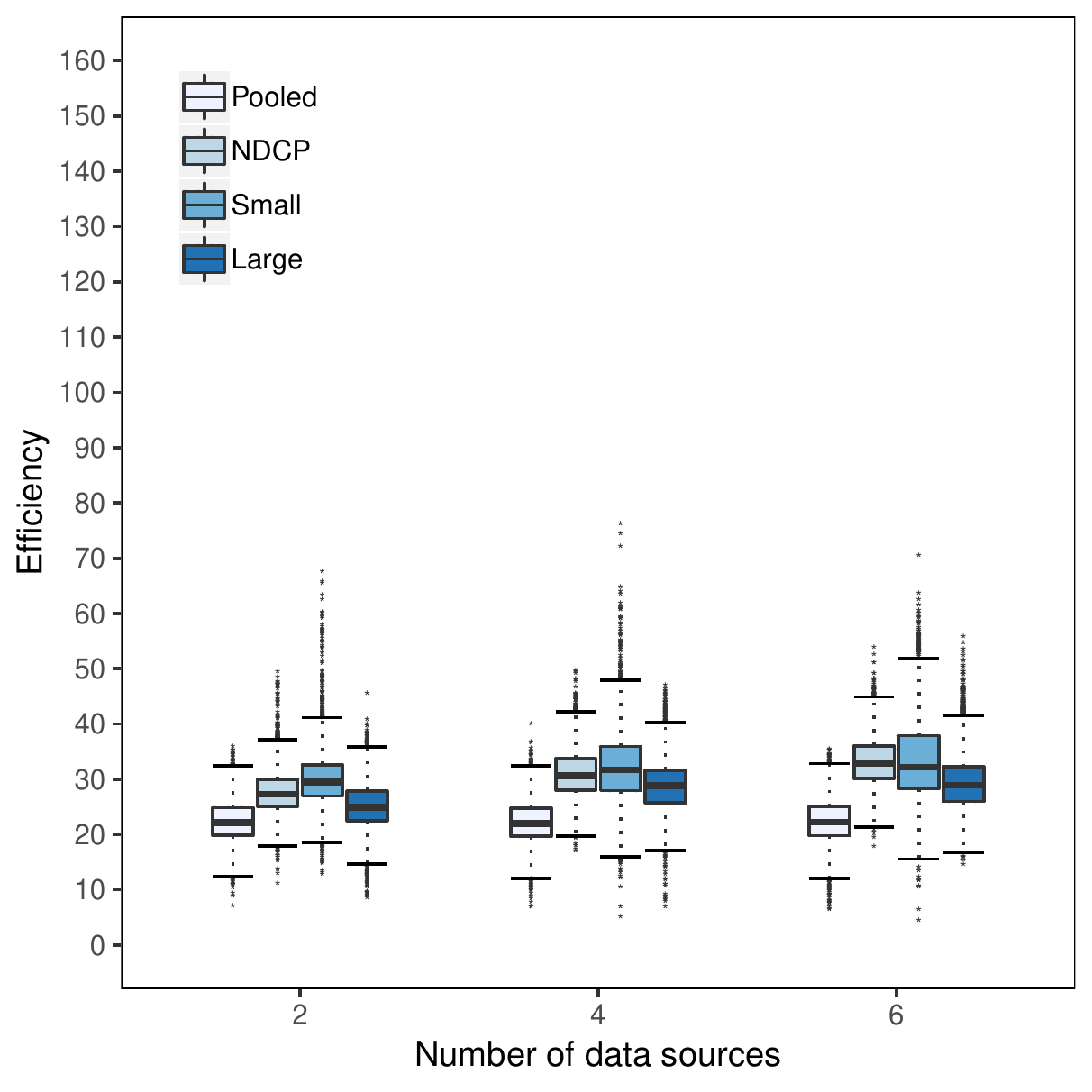}}
  }
   \caption{Dispersion of efficiency (prediction interval widths) for Experiment 2; unequally sized data sources, for 2, 4 and 6 number of data sources using ICP and CCP. Results from Pooled, NDCP and a randomly selected data source from the small data sources (Small), and the large data source (Large) are presented.}
  \label{fig:boxplot_size}
\end{figure}

In Figure \ref{fig:boxplot_size} the dispersion of the prediction interval widths are presented for Pooled, NDCP, and a randomly selected small data source and the large data source, where ICP and CCP is used for each model. We observe that while NDCP is capable of reducing the prediction interval variance compared to the model trained on the small sources, the model trained on the large data source still has a lower variance. This pattern is more clear with increasing number of data sources.


\begin{figure}[h]
  {%
    \subfigure[ICP]{\label{fig:boxplot_weight_icp}%
      \includegraphics[width=0.45\linewidth]{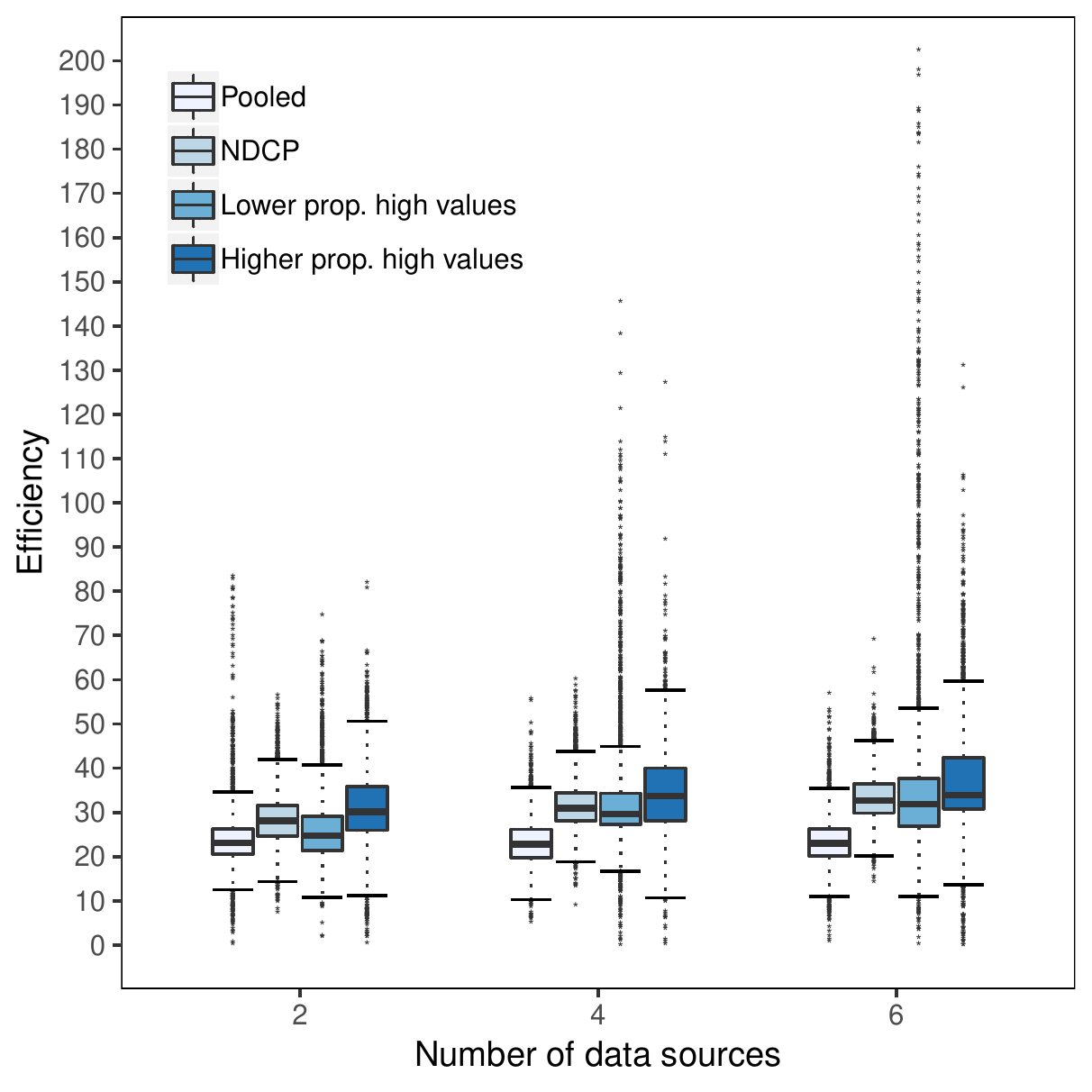}}%
    \qquad
    \subfigure[CCP]{\label{fig:boxplot_weight_ccp}%
      \includegraphics[width=0.45\linewidth]{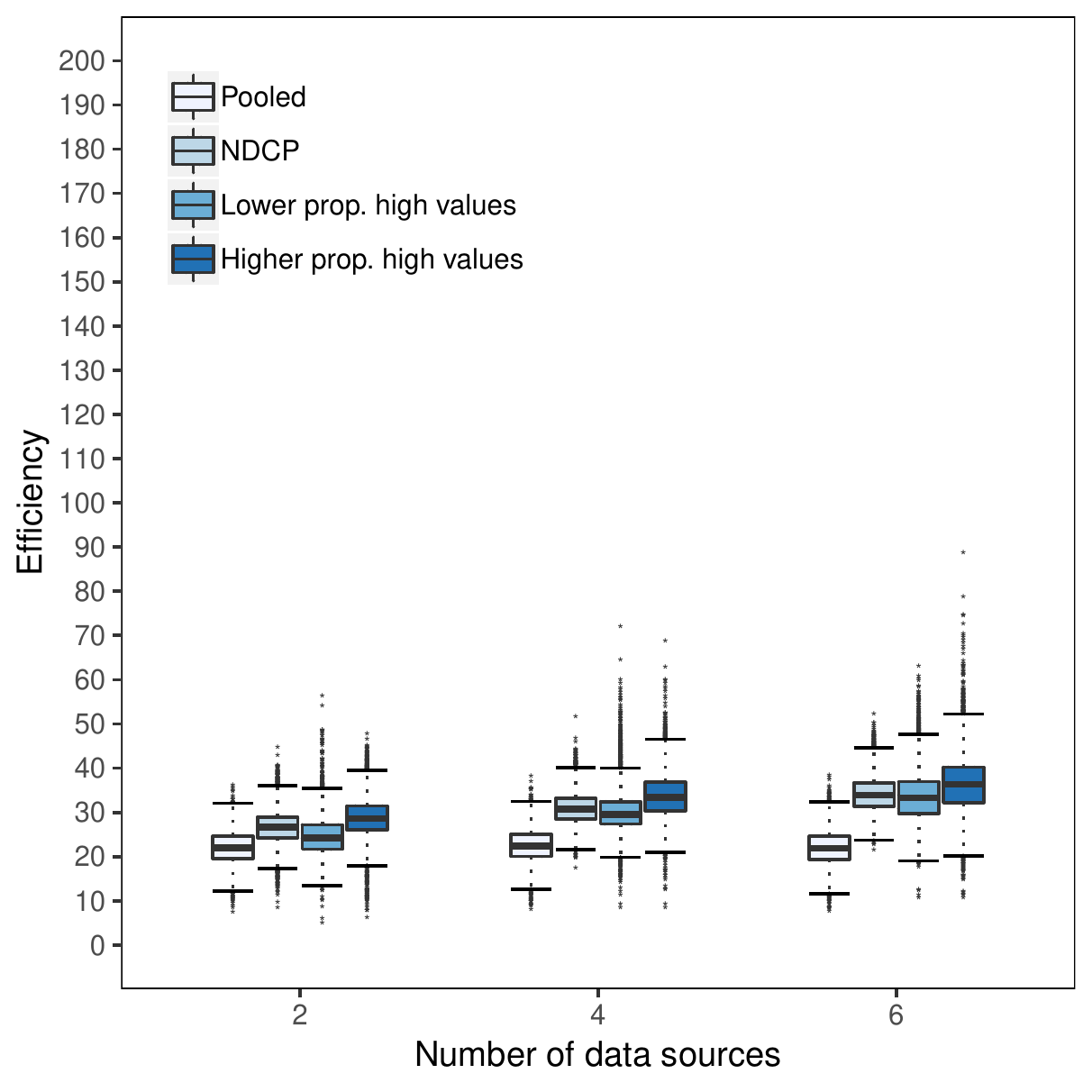}}
  }
   \caption{Dispersion of efficiency (prediction interval widths) for Experiment 3; Non-IID, equally sized data sources, for 2, 4 and 6 number of data sources using ICP and CCP. Results from Pooled, NDCP, a randomly selected data source with lower proportion of high-valued labels, and a data source with higher proportion of high-valued labels are presented.}
  \label{fig:boxplot_weight}
\end{figure}

\subsubsection{Experiment 3:  Non-IID, equally sized data sources}

In Table \ref{tab:res_weight} the results from splitting the data into non-IID, equally sized data sources are presented. In contrast to Experiment 1, data is distributed so that Source1 always contains a higher proportion of high-valued labels, which means that none of the sources will have identically distributed data compared to the test set. 

Also in this experiment we see that NDCP in all cases has a lower efficiency than at least one of the individual data sources, this is true for both ICP and CCP. For the three different scenarios, we observe large variance between the individual CCP and ICP sources where NDCP efficiency is consistently good, but not always the best. For the 2 data sources, we observe that Source2 has validity below 90\%. Source1 do however show an acceptable validity, but with a large interval. NDCP on the other hand manages to yield intervals with acceptable validity. Applying an ideal combination of intervals would give a tighter interval, which means there is room for improvement in the merging of intervals. For 4 and 6 sources we observe a similar pattern, and NDCP particularly improves in terms of validity when used with ICP.

\begin{table}
	\centering
	\footnotesize
	\caption{Results from Experiment 3, Non-IID, equally sized data sources, for models NDCP, Ideal NDCP, the individual equally sized data sources (2, 4 and 6) and Pooled. All models are listed in the first column and corresponding validity and PI median width as a measure of efficiency are listed in the next four columns for CCP and ICP respectively. $n$ refers to the number of observations underlying the predictions.} 
	\label{tab:res_weight}
	\begin{tabular}{lccccccc}
		\toprule
& \multicolumn{2}{c}{CCP} & \multicolumn{2}{c}{ICP} \\
		\textbf{2 sources}&  \textbf{Validity} &  \textbf{Efficiency} &  \textbf{Validity} & \textbf{Efficiency} & \textbf{n} \\ 
		\midrule
		NDCP & 0.959 & 26.510 & 0.953 & 27.790 & 927 \\ 
		Ideal NDCP &  0.950 & 25.816 & 0.950 & 27.381 & 927 \\ 
		Source1 &  0.946 & 28.878 & 0.936 & 30.602 & 463 \\ 
		Source2 &  0.895  & 24.281 & 0.879 & 25.055 & 464 \\ 
		Pooled & 0.960 & 22.137 & 0.943 &  22.968 & 927 \\ 
		\midrule
		\textbf{4 sources}&\\ 
		\midrule
		NDCP & 0.974 & 31.009 & 0.963 & 31.040 & 927 \\ 
		Ideal NDCP &  0.950 & 27.802 & 0.950 & 29.265 & 927 \\ 
		Source1 &  0.938 & 33.625 & 0.909 & 34.678 & 231 \\ 
		Source2 &  0.941 & 30.442 & 0.917 & 32.018 & 232 \\ 
		Source3 &  0.946 & 30.684 & 0.908 & 29.907 & 232 \\ 
		Source4 &  0.951 & 30.908 & 0.916 & 32.591 & 232 \\ 
		Pooled & 0.966 & 22.454 & 0.943 & 23.013 & 927 \\ 
		\midrule
		\textbf{6 sources}& \\
		\midrule
		NDCP &  0.974 & 33.723 & 0.965 & 32.825 & 927 \\ 
		Ideal NDCP &  0.950 & 29.819 & 0.950 & 30.552 & 927 \\ 
		Source1 &  0.928 & 37.158 & 0.887 & 37.143 & 154 \\ 
		Source2 &  0.944 & 33.401 & 0.907 & 37.042 & 155 \\ 
		Source3 &  0.947 & 34.521 & 0.916 & 33.508 & 154 \\ 
		Source4 &  0.942 & 33.450 & 0.920 & 35.658 & 155 \\ 
		Source5 &  0.945 & 34.148 & 0.912 & 38.356 & 155 \\ 
		Source6 & 0.944 & 33.387 & 0.906 & 32.591 & 154 \\ 
		Pooled &  0.958 & 22.091 & 0.947 & 23.494 & 927 \\ 
		\bottomrule
	\end{tabular}
\end{table}

In Figure \ref{fig:boxplot_weight} the dispersion of the interval widths for Pooled, NDCP, a data source with low proportion of high-valued labels and the data source with high proportion of high-valued labels are presented, where CCP is used for each model. For this simulated data partitioning, we observe a slightly tighter interval width for NDCP compared with the individual data sources, and in particular the data source with high proportion of high-valued labels. This result is more pronounced for the experiment with 6 data sources.


\section{Discussion}\label{sec:discussion}


This manuscript explores the use of combining prediction intervals from multiple conformal predictors in the case when data can't be disclosed between the individual data sources, and hence cannot be pooled into a traditional training set. In this scenario, it is also not disclosed the number of examples in each data source as this could be e.g. sensitive information. We performed a set of experiments to investigate our method Non-Disclosed Conformal Prediction (NDCP) for different data distribution scenarios between the individual data sources.

In all three experiments, NDCP do not perform as well as pooled data, but shows an improved efficiency over at least one of the individual data sources, which means it has some value for at least this data source. For equally sized data sources, NDCP compares very well and is mostly superior in terms of efficiency when compared to individual data sources. For unequally sized data sources the advantages of NDCP are less pronounced, but still NDCP outperforms at least one data source in all settings. For equally sized sources with non-IID, NDCP is consistently good, if not always the best, as compared with individual sources. For unequally sized sources, we could argue that the largest data source always outperforms NDCP; but in the NDCP setting the number of training objects is not disclosed so this will be hard to deduce without sharing potentially sensitive information. When the individual data sources do not have identical distributions compared to the test data, the individual data sources have larger variance in efficiency and generally lower validity, whereas NDCP presents models with good validity and good, if not always the best, models. We consider this scenario interesting as in real life scenarios the i.i.d. assumption is not always certain to fully hold. 

In general, when considering CCP vs ICP, aggregating seems to improve efficiency in our experiments although this is not in scope for this paper.

In this work we have done a relatively simple merging of intervals. Future work could include more advanced interval merging, such as weighting of intervals based on data source size (if such data can be disclosed). Considering the Ideal NDCP, which represents an optimal combination of intervals, shows that there indeed exist room of improvement in the merging of intervals.

The experiments with only two data sources is apparently a setting where NDCP is not as suitable. We also envision that NDCP would be yield improve results for larger numbers of data sources, which would be interesting to study in future experiments.

\section{Conclusions}

We present a method called Non-Disclosed Conformal Prediction (NDCP) to aggregate prediction intervals from multiple data sources while avoiding the pooling of data, thereby preserving data privacy. While we cannot retain the same efficiency as when all data is used, the efficiency is improved through the proposed approach as compared to predicting using a single source and in some evaluated scenarios is superior to models on all individual data sources. The results indicate that the NDCP method is relevant for predictions on non-disclosed data.


\section*{Acknowledgements}
This project received financial support from the Swedish Foundation for Strategic Research (SSF) as part of the HASTE project under the call `Big Data and Computational Science'. The computations were performed on resources provided by SNIC through Uppsala Multidisciplinary Center for Advanced Computational Science (UPPMAX) under project SNIC $2019/8-15.$  

\bibliography{ndcp}

\end{document}